\documentclass[11pt]{article} % For LaTeX2e
\usepackage{hyperref}
\usepackage{url}
\usepackage{smile}
\usepackage{graphicx} % more modern
\usepackage{algorithm}
\usepackage{algorithmic}

\usepackage{todonotes}
\usepackage{epstopdf}
\usepackage[margin=1in]{geometry}
\usepackage[normalem]{ulem}
\usepackage[export]{adjustbox}% http://ctan.org/pkg/adjustbox
\usepackage{mathtools, cuted}
\usepackage{natbib}
\usepackage{bbm}
\usepackage{wrapfig}
\usepackage{caption}
\usepackage{enumitem}

\makeatletter
\newenvironment{breakablealgorithm}
  {% \begin{breakablealgorithm}
   \begin{center}
     \refstepcounter{algorithm}% New algorithm
     \hrule height.8pt depth0pt \kern2pt% \@fs@pre for \@fs@ruled
     \renewcommand{\caption}[2][\relax]{% Make a new \caption
       {\raggedright\textbf{\ALG@name~\thealgorithm} ##2\par}%
       \ifx\relax##1\relax % #1 is \relax
         \addcontentsline{loa}{algorithm}{\protect\numberline{\thealgorithm}##2}%
       \else % #1 is not \relax
         \addcontentsline{loa}{algorithm}{\protect\numberline{\thealgorithm}##1}%
       \fi
       \kern2pt\hrule\kern2pt
     }
  }{% \end{breakablealgorithm}
     \kern2pt\hrule\relax% \@fs@post for \@fs@ruled
   \end{center}
  }
\makeatother

\hypersetup{
    colorlinks=true,
    linkcolor=blue,
    anchorcolor=blue, 
    citecolor=blue,
}

\usepackage{kpfonts}
\DeclareMathAlphabet{\mathsf}{OT1}{cmss}{m}{n}
\SetMathAlphabet{\mathsf}{bold}{OT1}{cmss}{bx}{n}

\providecommand{\norm}[1]{\|#1\|}

\begin{document}

\title{\huge {\bf Deep Reinforcement Learning with Robust and Smooth Policy}}

% It is OKAY to include author information, even for blind
% submissions: the style file will automatically remove it for you
% unless you've provided the [accepted] option to the icml2020
% package.

% List of affiliations: The first argument should be a (short)
% identifier you will use later to specify author affiliations
% Academic affiliations should list Department, University, City, Region, Country
% Industry affiliations should list Company, City, Region, Country

% You can specify symbols, otherwise they are numbe?red in order.
% Ideally, you should not use this facility. Affiliations will be numbered
% in order of appearance and this is the preferred way.
\setcounter{footnote}{1}

\author{
Qianli Shen{$^*$},  Yan Li{$^*$},  Haoming Jiang,  Zhaoran Wang, Tuo Zhao \footnote{Yan Li, Haoming Jiang and Tuo Zhao are affiliated with School of Industrial and Systems Engineering at Georgia Institute of Technology; Qianli Shen is affiliated with Yuanpei College at Peking University; Zhaoran Wang is affiliated with Department of 
Industrial Engineering and Management Sciences at Northwestern University; Tuo Zhao is the corresponding author; Email: tourzhao@gatech.edu. }
}
\date{\vspace{-5ex}}

\maketitle

% this must go after the closing bracket ] following \twocolumn[ ...

% This command actually creates the footnote in the first column
% listing the affiliations and the copyright notice.
% The command takes one argument, which is text to display at the start of the footnote.
% The \icmlEqualContribution command is standard text for equal contribution.
% Remove it (just {}) if you do not need this facility.

%\printAffiliationsAndNotice{}  % leave blank if no need to mention equal contribution

%!TEX root = ./SmoothRL.tex

\begin{abstract}
Deep reinforcement learning (RL) has achieved great empirical successes in various domains. 
However, the large search space of neural networks requires a large amount of data, which makes the current RL algorithms not sample efficient. 
Motivated by the fact that many environments with continuous state space have smooth transitions, we propose to learn a smooth policy that behaves smoothly with respect to states. 
We develop a new framework --- \textbf{S}mooth \textbf{R}egularized \textbf{R}einforcement \textbf{L}earning ($\textbf{SR}^2\textbf{L}$), where the policy is trained with smoothness-inducing regularization.
Such regularization effectively constrains the search space, and enforces smoothness in the learned policy. Moreover, our proposed framework can also improve the robustness of policy against measurement error in the state space, and can be naturally extended to distribubutionally robust setting. 
We apply the proposed framework to both on-policy (TRPO) and off-policy algorithm (DDPG). Through extensive experiments, we demonstrate that our method  achieves improved sample efficiency and robustness.

\end{abstract}
%!TEX root = ./SmoothRL.tex

\section{Introduction}

Deep reinforcement learning has enjoyed great empirical successes in various domains, including robotics, personalized recommendations, bidding, advertising and games \citep{levine2018learning,zheng2018drn,zhao2018deep,silver2017mastering,jin2018real}.
 At the backbone of its success is the superior approximation power of deep neural networks, which parameterize complex policy, value or state-action value functions, etc. 
 However, the high complexity of deep neural networks makes the search space of the learning algorithm prohibitively large, 
thus often requires a significant amount of training data, and suffers from numerous training difficulties such as overfitting and training instability \citep{thrun1993issues,boyan1995generalization,zhang2016understanding}.

Reducing the size of the search space while maintaining the network's performance requires special treatment. 
While one can simply switch to a network of smaller size, numerous empirical evidences have shown that small network often leads to performance degradation and training difficulties. 
It is widely believed that  training a sufficiently large network (also known as over-parameterization) 
with suitable regularization (e.g., dropout \cite{srivastava2014dropout}, orthogonality parameter constraints \cite{huang2018orthogonal}) is the most effective way to adaptively constrain the search space, while maintaining the performance of a large network.

For reinforcement learning problems, 
entropy regularization is one commonly adopted regularization, which is believed to help facilitate exploration in the learning process.
Yet in the presence of high uncertainty in the environment and large noise, such regularization might yield poor performances.
More recently, \citet{pinto2017robust} propose robust adversarial reinforcement learning (RARL) that 
aims to perform well under uncertainties by training the agent to be robust against the adversarially perturbed environment. 
However, in addition to the marginal performance gain,  the requirement of learning the additional adversarial policy makes the update of RARL computationally expensive and less sample efficient than traditional learning algorithms.
 \citet{cheng2019control} on the other hand propose a control regularization that enforces the behavior of the deep policy to be similar to a policy prior, yet designing a good prior often requires a significant amount of domain knowledge. 

Different from previous works, we propose a new training framework -- \textbf{S}moothness \textbf{R}egularized \textbf{R}einforcement \textbf{L}earning ($\textbf{SR}^2\textbf{L}$) for training reinforcement algorithms.
Through promoting smoothness, we effectively reduce the size of the search space when learning the policy network and achieve state-of-the-art sample efficiency.
Our goal of promoting smoothness in the policy is motivated by the fact that natural environments with continuous state space often have smooth transitions from state to state, which favors a smooth policy -- similar states leading to similar actions. 
As a concrete example, for  MuJoCo environment  \citep{todorov2012mujoco}, which is a system powered by physical laws, 
the optimal policy can be described by a set of differential equations with certain smoothness properties.

Promoting smoothness is particularly important for deep RL, since deep neural networks can be extremely non-smooth, due to their high complexity.
 It is observed that small changes in neural networks' input would result in significant changes in its output \citep{goodfellow2014explaining,kurakin2016adversarial}.
%Such non-smoothness has drawn significant attention in other domains (e.g., image recognition, information security) that involve using neural networks as a part of the decision process \citep{goodfellow2014explaining,kurakin2016adversarial}.
To train a smooth neural network, we need to employ many hacks in the training process.
In supervised learning setting with i.i.d. data, these hacks include but not limited to batch normalization \citep{ioffe2015batch}, layer normalization \citep{ba2016layer}, orthogonal regularization \citep{huang2018orthogonal}.
However, most of existing hacks do not work well in RL setting, where the training data has complex dependencies.
As one significant consequence,  current reinforcement learning algorithms often lead to undesirable non-smooth policy.

Our proposed $\textbf{SR}^2\textbf{L}$ training framework uses a smoothness-inducing regularization to encourage the output of the policy (decision) to not change much when injecting small perturbation to the input of the policy (observed state).
The  framework is motivated by  local shift sensitivity in robust statistics literature \citep{hampel1974influence}, which can also be considered as a measure of the local Lipschitz constant of the policy.
We highlight that  $\textbf{SR}^2\textbf{L}$ is highly flexible and can be readily adopted  into  various reinforcement learning algorithms.
As concrete examples, we apply $\textbf{SR}^2\textbf{L}$ to the TRPO algorithm \citep{schulman2015trust}, which is an \textit{on-policy} method, and the regularizer directly penalizes non-smoothness of the policy.
In addition, we also apply $\textbf{SR}^2\textbf{L}$ to DDPG algorithm \citep{lillicrap2015continuous}, which is an \textit{off-policy} method, and the regularizer penalizes non-smoothness of either the policy or the state-action value function (also known as the Q-function), which can be used to induce a smooth policy. 

Moreover, we remark that besides promoting the smoothness of policy network, our proposed regularizer can also help improve the robustness of policy against random, or even adversarial measurement error in the state space.

Our proposed smoothness-inducing regularizer is related to several existing works \citep{miyato2018virtual,zhang2019theoretically,hendrycks2019using,xie2019unsupervised,jiang2019smart}. These works consider similar regularization techniques, but target at other applications with different motivations, e.g., semi-supervised learning, unsupervised domain adaptation and harnessing adversarial examples.

The rest of the paper is organized as follows: Section 2 introduces the related background; Section 3 introduces our proposed smooth regularized reinforcement learning ($\textbf{SR}^2\textbf{L}$) in detail; Section 4 discusses how our proposed method help improve the robustness and a natural extension to distributionally robust settings; Section 5
presents numerical experiments on various MuJoCo environments to demonstrate the superior performance of $\textbf{SR}^2\textbf{L}$; Section 6 draws a brief conclusion.

%\textbf{Notations:}
%We let $\mathbb{B}_d(s,\epsilon) = \{\tilde{s}: d(s, \tilde{s}) \leq \epsilon \}$ denote the $\epsilon$-radius ball measured in metric $d(\cdot, \cdot)$ centered at point $s$.
%We use $I_A\in \RR^A$ to denote the identity matrix in $A$-dimensional Euclidean space. 

%!TEX root = ./SmoothRL.tex
%\vspace{-0.1in}
\section{Background}
We consider a Markov Decision Process  $(\cS, \cA, \PP, r, p_0, \gamma)$, in which an agent interacts with an environment in discrete time steps.
We let 
 $\cS \subseteq \RR^S$ denote the continuous state space,
 $\cA \subseteq \RR^A$ denote the action space,
$\PP: \cS \times \cA \rightarrow \cS$ denote the transition kernel,
 $r: \cS \times \cA \rightarrow \RR$ denote the reward function,
$p_0$ denote the initial distribution and $\gamma$ denote the discount factor.
An agent's behavior is defined by a policy, either stochastic or deterministic. A stochastic policy $\pi$ maps a state to a probability distribution over the action space $\pi: \cS \rightarrow \cP (\cA)$. 
A deterministic policy $\mu$ maps a state directly to an action $\mu: \cS \rightarrow \cA$.
At each time step, the agent observes its state $s_t \in \cS$, takes action $a_t \sim \pi(s_t)$, and receives reward $r_t = r(s_t, a_t)$.
The agent then transits into the next state $s_{t+1}$ following the transition kernel $s_{t+1} \sim \PP(\cdot | s_t, a_t)$.
The goal of the agent is to find a policy that maximize the expected discounted reward:
%\vspace{-0.1in}
\begin{align*}
\max_{\pi} V (\pi)   = \EE_{s_0, a_0, \ldots}   \Bigg[\sum_{t\geq 0} \gamma^t r (s_t, a_t)\Bigg], 
\end{align*}
 with $  s_0  \sim p_0, a_t \sim \pi(s_t), s_{t+1}\sim \PP(s_{t+1} | s_t, a_t).$
One way to solve the above problem is the classical policy gradient algorithms,  which estimate the gradient of the expected reward through trajectory samples, and update the parameters of the policy by following the estimated gradient.
The policy gradient algorithms suffers from high variance of estimated gradient, which often leads to aggressive updates and unstable training. 
To address this issue, numerous variants have been proposed. Below we briefly review two  popular ones used in practice. 

\subsection{Trust Region Policy Optimization (TRPO)}

TRPO iteratively improves a parameterized  policy $\pi_\theta$ by solving a trust region type optimization problem.
Before we describe the algorithm in detail, we need several definitions in place.
The value function $V^\pi(s)$ and the state-action value function $Q^\pi(s,a)$ are defined by:
\vspace{-0.1in}
\begin{align*}
V^\pi  (s)    = \EE_{s_0 = s, a_0, \ldots}   \Bigg[\sum_{t\geq 0} \gamma^t r (s_t, a_t)\Bigg], ~~
	Q^{\pi} (s,a)  = \EE_{s_1, a_1, \ldots} \Bigg[\sum_{t\geq 0} \gamma^t r (s_t, a_t) \vert s_0 = s, a_0 = a\Bigg], 
\end{align*}
with $ a_t \sim \pi(s_t), s_{t+1}\sim \PP(s_{t+1} | s_t, a_t).$
The advantage  function $A^\pi(s,a)$ and the discounted state visitation distribution (unnormalized) $\rho^\pi(s) $ are defined by:
\begin{align*}
A^{\pi} (s,a) = Q^{\pi} (s,a) - V^{\pi} (s), ~~ \rho^\pi(s) = \sum_{i\geq 0}\gamma^i \PP(s_i=s).
\end{align*}
At the $k$-th iteration of TRPO, the policy is updated by:
\begin{align} \label{eq:trpo}
	\theta_{k+1} = \argmax_{\theta}  ~~  \EE_{s \sim \rho^{\pi_{\theta_{k} }}, \atop a \sim \pi_{\theta_{k} }} \left[ \frac{\pi_{\theta}(a|s)}{\pi_{\theta_{k}}(a|s) } A^{\pi_{\theta_{k}}} (s,a) \right], 
	~~\textrm{subject to}~  \EE_{s \sim \rho^{\pi_{\theta_{k}}}}  \left[ \cD_{\textrm{KL}}( \pi_{\theta_{k}}(\cdot|s) \| \pi_{\theta}(\cdot|s_ ) \right] \leq \delta ,
\end{align}
where $\delta$ is a tuning parameter for controlling the size of the trust region, and 
$\cD_\mathrm{KL}(P||Q) = \int_{\mathcal{X}} \log \rbr{\frac{ dP }{dQ}} dQ$ denotes the  Kullback-Leibler divergence between two distributions $P, Q$ over support $\mathcal{X}$.
For each update, the algorithm: (i) Samples trajectories using the current policy $\pi_{\theta_{k}}$; (ii) Approximates $A^{\pi_{\theta_{k}}} $ for each state-action pair by taking the discounted sum of future rewards along the trajectory; (iii) Replaces the expectation in \eqref{eq:trpo} and $A^{\pi_{\theta_{k}}} $  by sample approximation, then solves  \eqref{eq:trpo} with conjugate gradient algorithm.

\subsection{Deep Deterministic Policy Gradient (DDPG)}
DDPG uses the actor-critic architecture, where the agent learns a parameterized state-action value function $Q_{\phi}$ (also known as the critic) to update the parameterized deterministic policy $\mu_\theta$ (also known as the actor).

DDPG uses a replay buffer, which is also used in  in Deep Q-Network \cite{mnih2013playing}.
The replay buffer  is a finite sized cache.
 Transitions are sampled from the environment according to the policy and the tuple
$(s_t, a_t, r_t, s_{t+1})$ is stored in the replay buffer. When the replay buffer is full, the oldest samples are discarded. At each step, $\mu_{\theta}$ and $Q_{\phi}$ are updated by sampling a mini-batch from the buffer.

\textbf{Update of state-action value function. } 
The update of the state-action value function network $Q_\phi$ depends on the deterministic Bellman equation:
\begin{align}\label{eq:deter_bellman}
Q_{\phi} (s,a) = \EE_{  s' \sim \PP(\cdot |s,a)} \left[r(s,a) + \gamma Q_{\phi} (s',\pi_\theta(s')) \right].
\end{align}
The expectation depends only on the environment. 
This means that unlike TRPO, DDPG is an off-policy method, which can use transitions generated from a different stochastic behavior policy denoted as $\beta$ (see \citet{lillicrap2015continuous} for detail).
At the $t$-th iteration, we update the $Q_\phi$  by minimizing the associated mean squared Bellman error of transitions $\{(s^i_t, a^i_t, r^i_t, s^i_{t+1})\}_{i \in B}$ sampled  from the replay buffer.
Specifically, let $Q_{\phi_t'}$ and $\mu_{\theta_t'}$ be a pair of target networks,  we set $y^i_t = r^i_t + \gamma Q_{\phi_t'}(s^i_{t+1}, \mu_{\theta_t'}(s^i_{t+1}))$, and then update the critic network:
$\phi_{t+1} = \argmin_{\phi} \sum_{i \in B} \rbr{ y^i_t - Q_\phi ( s^i_t, a^i_t)}^2.
$ After both critic and actor networks are updated, we update the target networks by slowly tracking the critic and actor networks:
$\phi_{t+1}' = \tau \phi_{t+1} + (1-\tau) \phi_{t}', \theta_{t+1}' = \theta \phi_{t+1} + (1-\tau) \theta_{t}'
$
with, $\tau \ll 1$.

\textbf{Update of policy. } The policy network $\mu_\theta$ is updated by maximizing the value function using policy gradient:
\begin{align}\label{eq:ddpg_policy_obj}
\max_{\theta} \mathop{\EE}_{s \sim \rho^\beta}\left[ Q_{\phi} (s,a)\big|_{a=\pi_\theta(s)} \right].
\end{align}
Similar to updating the critic, we use the minibatch sampled from the replay buffer to compute approximated gradient of $\theta$ and perform the update:
$\theta_{t+1} = \theta_t + \frac{\eta_t}{|B|} \sum_{i \in B} \nabla_a Q_{\phi_{t+1}} (s,a)\Big|_{s= s_i, \atop a = \mu_{\theta_t}(s_i)} \nabla_{\theta} \mu_{\theta_t}(s_i).
$

%!TEX root = ./SmoothRL.tex

\section{Method}
We present the smoothness-inducing regularizer  in its general form and describe its intuition in detail.
We also apply the proposed regularizer to popular reinforcement learning algorithms  to demonstrate its great adaptability. 

\subsection{Learning Policy with $\textbf{SR}^2\textbf{L}$}

 We first focus on directly learning smooth policy $\pi_\theta$ with the proposed regularizer. 
We assume that the state space is continuous, i.e., $\cS \subseteq \RR^S$. 

For a fixed state $s \in \cS$ and a policy $\pi_\theta$,  $\textbf{SR}^2\textbf{L}$ encourages the output  $\pi_\theta(s)$ and $\pi_\theta(\tilde{s})$ to be similar, where state  $\tilde{s}$ is obtained by injecting a small perturbation to state $s$.
%We let $\mathbb{B}_d(s,\epsilon) = \{\tilde{s}: d(s, \tilde{s}) \leq \epsilon \}$ denote the $\epsilon$-radius ball measured in metric $d(\cdot, \cdot)$ centered at point $s$.
We assume the perturbation set $\mathbb{B}_d(s,\epsilon) = \{\tilde{s}: d(s, \tilde{s}) \leq \epsilon \}$ is an $\epsilon$-radius ball measured in metric $d(\cdot, \cdot)$, which is often chosen to be the $\ell_p$ distance: $d(s,\tilde{s}) = \norm{s - \tilde{s}}_p$.
To measure the discrepancy between the outputs of a policy, we adopt a suitable metric function denoted by $\cD$. 
The non-smoothness of policy $\pi_\theta$ at state $s$ is defined in an adversarial manner:
$  \max_{\tilde{s}\in 
\mathbb{B}_d(s,\epsilon)}\cD(\pi_\theta(s),\pi_\theta(\tilde{s})).
$
To obtain a smooth policy $\pi_\theta$, we encourage smoothness at each state of the entire trajectory.
We achieve this by taking expectation with respect to the state visitation distribution $\rho^\pi$ induced by the policy,
and our smoothness-inducing regularizer is defined by:
\begin{align}\label{eq:regularizer}
\cR_\mathrm{s}^\pi(\theta)=  \mathop{\EE}_{s\sim\rho^{\pi_{\theta_t}}} \max_{\tilde{s}\in 
\mathbb{B}_d(s,\epsilon)}\cD(\pi_\theta(s),\pi_\theta(\tilde{s})).
\end{align}
For a stochastic policy $\pi_\theta$, we set the  metric $\cD(\cdot, \cdot)$ to be the Jeffrey's divergence, and the regularizer takes the form
\begin{align}\label{eq:SR_stochastic_KLdiv} 
\cR_{\rm s}^\pi(\theta) = \mathop{\mathbb{E}}_{s\sim\rho^\pi}\max_{\tilde{s}\in \mathbb{B}_d(s,\epsilon)}\cD_{\mathrm{J}}(\pi(\cdot|s)\ ||\ \pi(\cdot|\tilde{s})),
\end{align}
where the Jeffrey's divergence for two distributions $P, Q$ is defined by:
\begin{align}
\cD_{\mathrm{J}} (P || Q) = \frac{1}{2} \cD_{\mathrm{KL}}(P||Q) + \frac{1}{2} \cD_{\mathrm{KL}}(Q||P).
\end{align}
For a deterministic policy $\mu_\theta$, we set the metric  $\cD(\cdot, \cdot)$ to be the squared $\ell_2$ norm of the difference:
\begin{align}\label{eq:SR_deterministic_sqd} 
\cR_{\rm s}^\mu(\theta) = \mathop{\mathbb{E}}_{s\sim \rho^\mu}\max_{\tilde{s}\in \mathbb{B}_d(s,\epsilon)} \norm{\mu(s)-\mu(\tilde{s})}_2^2.
\end{align}
\begin{figure}[!b]
	\centering
	\includegraphics[width = 0.45\textwidth]{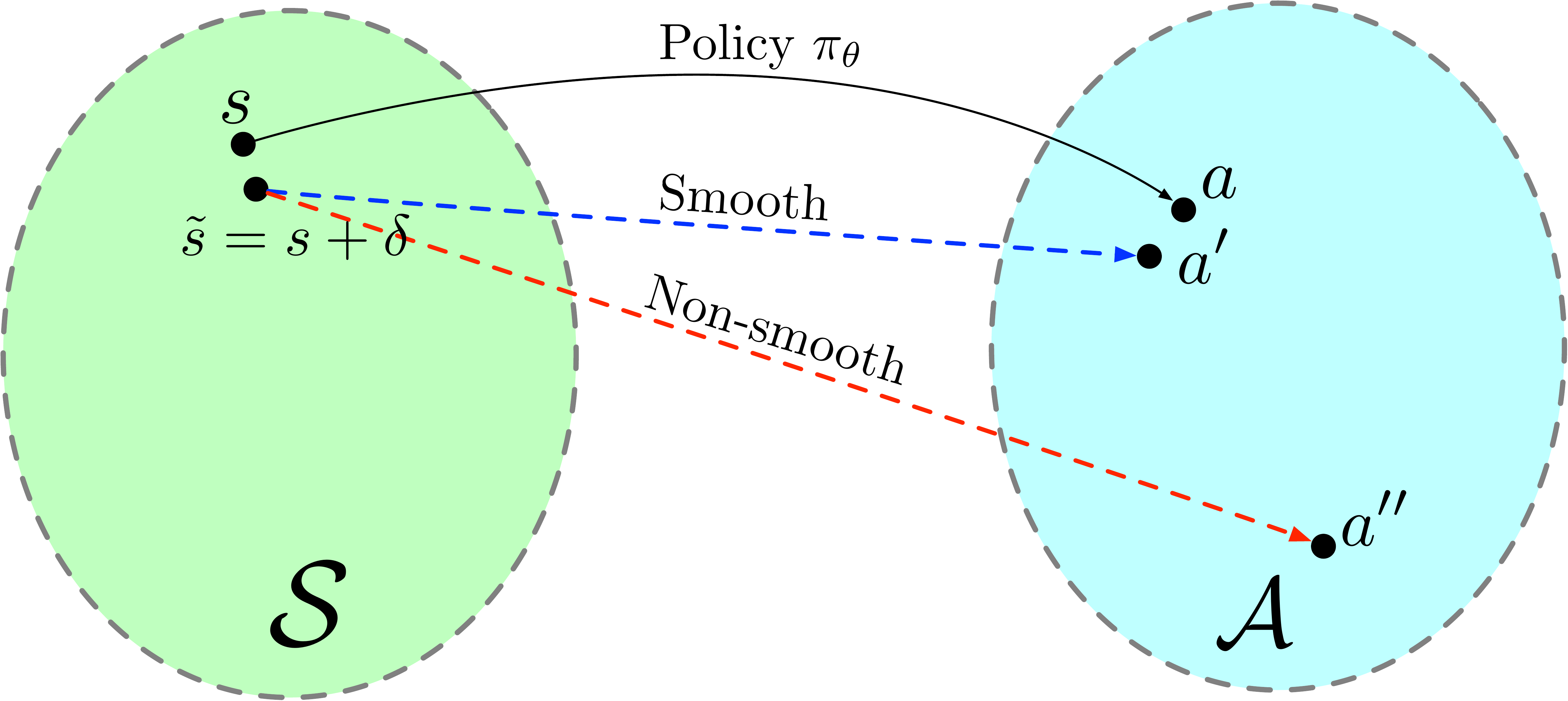}
	\vspace{-0.1in}
	\caption{Smoothness of policy $\pi_\theta$ at state $s$. If policy $\pi_\theta$ is smooth at state $s$, then perturbed state $\tilde{s}$ leads to action $a'$ similar to the original action $a$. If the policy $\pi_\theta$ is non-smooth at state $s$, then the perturbed state $\tilde{s}$ leads to drastically different action $a''$. }
	\label{fig:policy}
\end{figure}
The smoothness-inducing adversarial regularizer is essentially measuring the local Lipschitz continuity of policy  $\pi_\theta$ under the metric $\cD$. More precisely, we encourage the output (decision) of $\pi_\theta$ to not change much if we inject a small perturbation  bounded in metric $d(\cdot, \cdot)$ to the state $s$ (See Figure \ref{fig:policy}).
Therefore, by adding the regularizer \eqref{eq:regularizer} to the policy update, 
we can encourage $\pi_\theta$ to be smooth within the neighborhoods of all states on all possible trajectories regarding to the sampling policy. Such a smoothness-inducing property is particularly helpful to prevent overfitting, improve sample efficiency and overall training stability .

\textbf{TRPO with $\textbf{SR}^2\textbf{L}$ }(TRPO-SR). 
We now apply the proposed smoothness inducing regularizer to TRPO algorithm, which is itself an on-policy algorithm.
Since TRPO uses a stochastic policy, we use the Jeffrey's divergence to penalize the discrepancy between decisions for the regular state and the adversarially perturbed state, as suggested in \eqref{eq:SR_stochastic_KLdiv}.

Specifically, TRPO with smoothness-inducing regularizer updates the policy by solving the following subproblem at the $k$-th iteration:
\begin{align}
	\theta_{k+1}  & = \argmin_{\theta}   -\EE_{s \sim \rho^{\pi_{\theta_{k} }}, \atop a \sim \pi_{\theta_{k} }}  \left[ \frac{\pi_{\theta}(a|s)}{\pi_{\theta_{k}}(a|s) }  A^{\pi_{\theta_{k}}} (s,a) \right] 
 \quad +   \lambda_{\rm s} \EE_{s \sim \rho^{\pi_{\theta_{k} }}}  \max_{\tilde{s}\in \mathbb{B}_d(s,\epsilon)}   \cD_{\mathrm{J}}(\pi_{\theta}(\cdot|s)\ ||\ \pi_{\theta}(\cdot|\tilde{s})), \nonumber \\
	\text{s.t.} &~~~~ \EE_{s \sim \rho^{\pi_{\theta_{k}}}}   \left[ \cD_{\textrm{KL}} ( \pi_{\theta_{k}}(\cdot|s) \| \pi_{\theta}(\cdot|s_ ) \right] \leq \delta.
\end{align}

\subsection{Learning Q-function with Smoothness-inducing Regularization}
The proposed smoothness-induced regularizer can be also used to learn a smooth Q-function, which can be further used to generate a smooth policy.

We measure the non-smoothness of a Q-function at state-action pair $(s,a)$ by the squared difference of the state-action value between the normal state and the adversarially perturbed state:
$\max_{\tilde{s}\in \mathbb{B}_d(s,\epsilon)}(Q_{\phi}(s,a)-Q_{\phi}(\tilde{s},a))^2.
$
To enforce smoothness at every state-action pair, we take expectation with respect to the entire trajectory, and the smoothness-inducing regularizer takes the form
\begin{align*}
\cR^Q_s(\phi) = \mathbb{E}_{s\sim\rho^{\beta},\atop a\sim\beta}\max_{\tilde{s}\in \mathbb{B}_d(s,\epsilon)}(Q_{\phi}(s,a)-Q_{\phi}(\tilde{s},a))^2.
\end{align*}
where $\beta$ denotes the behavior policy for sampling in off-policy training setting.

\textbf{DDPG with $\textbf{SR}^2\textbf{L}$}.
We now apply the proposed smoothness-inducing regularizer to DDPG algorithm, which is itself an off-policy algorithm.
Since DDPG uses two networks: the actor network and the critic network,
we propose two variants of DDPG, where the regularizer is  applied to the actor or the critic network.

$\bullet$ \textbf{Regularizing the Actor Network} (DDPG-SR-A).
We can directly penalize the non-smoothness of the actor network to promote a smooth policy in DDPG.
Since DDPG uses a deterministic policy $\mu_\theta$, when updating the actor network, we  penalize the squared difference as suggested in \eqref{eq:SR_deterministic_sqd} and minimize the following objective:
\begin{align*}
 \mathop{\EE}_{s \sim \rho^\beta}\left[ -Q_{\phi} (s,a)\big|_{a=\mu_{\theta}(s)}  +  \lambda_{\rm s} \max_{\tilde{s}\sim \mathbb{B}_d(s,\epsilon)}\norm{\mu_{\theta}(s)-\mu_{\theta}(\tilde{s})}_2^2  \right].
\end{align*}
The policy gradient can be written as:
$\mathop{\EE}_{s \sim \rho^\beta}  \Big[ -  \nabla_{a} Q_{\phi} (s,a)\big|_{a=\mu_{\theta}(s)}   \nabla_{\theta}  \mu_{\theta}(s)   + \lambda_{\rm s} \nabla_{\theta}\norm{\mu_{\theta}(s)-\mu_{\theta}(\tilde{s})}_2^2   \Big],$
  with $\displaystyle \tilde{s}=\argmax_{\tilde{s}\sim \mathbb{B}_d(s,\epsilon)} \norm{\mu_{\theta}(s)-\mu_{\theta}(\tilde{s})}_2^2$  for $s \sim \rho^\beta$.

$\bullet$ \textbf{Regularizing the Critic Network} (DDPG-SR-C).
Since DDPG simultaneously learns a  Q-function (critic network) to update the policy (actor network), inducing smoothness in the critic network could also help us to generate a smooth policy. 
By incorporating the proposed regularizer for penalizing Q-function, we obtain the following update for inducing a smooth Q-function in DDPG:
\begin{align*}
\phi_{t+1}  = \argmin_{\phi}   \sum_{i \in B} \rbr{ y^i_t - Q_\phi ( s^i_t, a^i_t)}^2 
 \quad + \lambda_{\rm s}\sum_{i \in B} \max_{\tilde{s_t^i}\sim \mathbb{B}_d(s_t^i,\epsilon)}(Q_{\phi}(s_t^i,a_t^i)-Q_{\phi}(\tilde{s}_t^i,a_t^i))^2 ,
 \end{align*}
with $y^i_t = r^i_t + \gamma Q_{\phi_t'}(s^i_{t+1}, \mu_{\theta_t'}(s^i_{t+1})), \forall i \in B$,
where $B$ is the mini-batch sampled from the replay buffer.

\subsection{Solving the Min-max Problem}
Adding the smoothness-inducing regularizer in the policy/Q-function update often involves solving a min-max problem. Though the inner max problem is not concave, simple stochastic gradient algorithm has been shown to be able to solve it efficiently in practice.
Below we describe how to perform the update of TRPO-SR, including solving the corresponding min-max problem.
The details are summarized in Algorithm \ref{alg:trpo-sr}. 
We leave the detailed description of DDPG-SR-A and DDPG-SR-C in the appendix.
\begin{algorithm}[tbh!]
    \caption{Trust Region Policy Optimization with Smoothness-inducing Regularization.}
    \label{alg:trpo-sr}
    \begin{algorithmic}
    	\STATE{\textbf{Input}: step sizes $\eta_\delta$, $\eta_\theta$, number of iterations $D$ for inner optimization, number of iterations $K$ for policy updates, perturbation strength $\epsilon$, regularization coefficient $\lambda_\mathrm{s}$.}
	\STATE{\textbf{Initialize}: randomly initialize the policy network $\pi_{\theta_0}$.}
	\FOR{$k=1,\ldots, K-1$}
	\STATE{Sample trajectory $\cS_k = \{(s_k^t, a_k^t)\}_{t=1}^T$ from current policy $\pi_{\theta_k}$.}
	\STATE{Estimate advantage function $\hat{A}^{\pi_{\theta_k}}(s,a)$ using sample approximation.}
	\FOR{$s^t_k \in \cS_k$}
	\STATE{Randomly initialize $\delta^t_0$.}
	\FOR{$\ell = 0, \ldots, D-1$}
	\STATE{$\delta_{\ell+1}^t = \delta_\ell^t + \eta_\delta \nabla_\delta \cD_{\mathrm{JS}} (\pi_{\theta_k}(\cdot|s^t_k ) || \pi_{\theta_k}(\cdot|s^t_k + \delta^t_\ell ) ) $.}
	\STATE{$\delta_{\ell+1}^t =  \Pi_{\mathbb{B}_d(0, \epsilon)} \left(\delta_{\ell+1}^t \right) $.}
	\ENDFOR
	\STATE{$\tilde{s}^t_k = s^t_k + \delta^t_D$.}
	\ENDFOR
	\vspace{-0.3 in}
	\STATE{\begin{align*}\theta_{k+1}  =  \theta_k  + \eta_\theta \sum_{(s^t_k, a^t_k) \in \cS_k} \frac{\hat{A}^{\pi_{\theta_k}}(s_k^t, a_k^t)}{\pi_{\theta_k}(a_k^t|s_k^t)} \nabla_\theta \pi_{\theta_k} (a_k^t|s_k^t)   - \eta_\theta \lambda_\mathrm{s}  \sum_{s_k^t \in \cS_k} \gamma^t \nabla_\theta  \cD_{\mathrm{J}} (\pi_{\theta^k}(\cdot|s_k^t ) || \pi_{\theta^k}(\cdot|\tilde{s}^t_k) ).\end{align*}}
	\vspace{-0.2 in}
	\ENDFOR
    \end{algorithmic}
\end{algorithm}
    	\vspace{-0.2 in}

\section{Connection to (Distributionally) Robust Reinforcement Learning}

%\textbf{Robustness to Measurement Error}.
%While in a simulated environment, one can provide the learning agent the exact state information, in practice this is often not the case.
Besides promoting the smoothness of the learnt policy, our proposed regularizer also enjoys  another advantage -- improving {\it robustness against measurement error} in the state space. Specifically, we consider a noisy reinforcement learning enviroment, where the agent can only observe inexact state information.
Taking the robot motion planning as an example, the robot gets its locations and velocity from the equipped sensors, which often encouter systematic or stochastic measurement error.

To address this issue, researchers usually resort to Partially observable Markov decision process (POMDP, \citet{monahan1982state}) to model the inexact state information as a conditional distribution that depends on the exact state. However, POMDP can only handle i.i.d. stochastic measurement errors, and require the prior knowledge of the measurement error distribution.
In contrast, our regularizer can handle more complex non i.i.d. error, or even adversarial measurement error. Our proposed regularizer encourages the policy to make similar actions for any pair of states that are close to each other,  which implies that actions for the observed state and true state should be close. 
This inductive bias is well suited for RL with smooth environments, whose optimal policy does not drastically change its decision when adding a small perturbation to the state.

Our regularizer can be naturally extended to {\it distributionally robust} settings, where the observed state comes from a state-visitation distribution $\rho'$ that is close to the state-visitation distribution $\rho$ of the true state.
%where the 
%goal is to learn a policy that is robust against measurement error $\delta$ with distribution $\PP_\delta$ coming from a set of specified distribution set $\cP$.
Specifically, the regularizer takes the form $$\cR_\mathrm{s}^\pi(\theta) = \max_{\cF(\rho, \rho') \leq \epsilon} \EE_{s \sim \rho, s' \sim \rho'} \cD\rbr{\pi_{\theta}(s), \pi_\theta(s')},$$
where $\cF(\cdot, \cdot)$ denotes some discrepancy measure between a pair of visitation probability distributions (e.g., Wasserstein distance, $f$-divergence). For more details on solving distributionally robust optimization via duality, please refer to \citet{gao2016distributionally}.

%\begin{align*}
%\max_{\PP_\delta \in \cP} \EE_{\delta \sim \PP_\delta} \cD\rbr{\pi_{\theta}(s), \pi_\theta(s+\delta)}.
%\end{align*}
% The distributionally robust setting generalizes previous random/adversarial  measurement error setting as special case. 
% Specifically, for $\cP = \{\mathrm{Unif}(\mathbb{B}_d(0,\epsilon)\}$, we recover the random measurement error setting.
% For $\cP = \cup_{\delta \in \mathbb{B}_d(0,\epsilon)} \{\mathbbm{1}_{\delta} \}$, where $\mathbbm{1}_{\delta} $ denotes the Dirac measure
%centered at $\delta$, we recover the adversarial measurement error setting.

%!TEX root = ./SmoothRL.tex

 \begin{figure*}[!b]
	\centering
	\includegraphics[width = 0.45\textwidth]{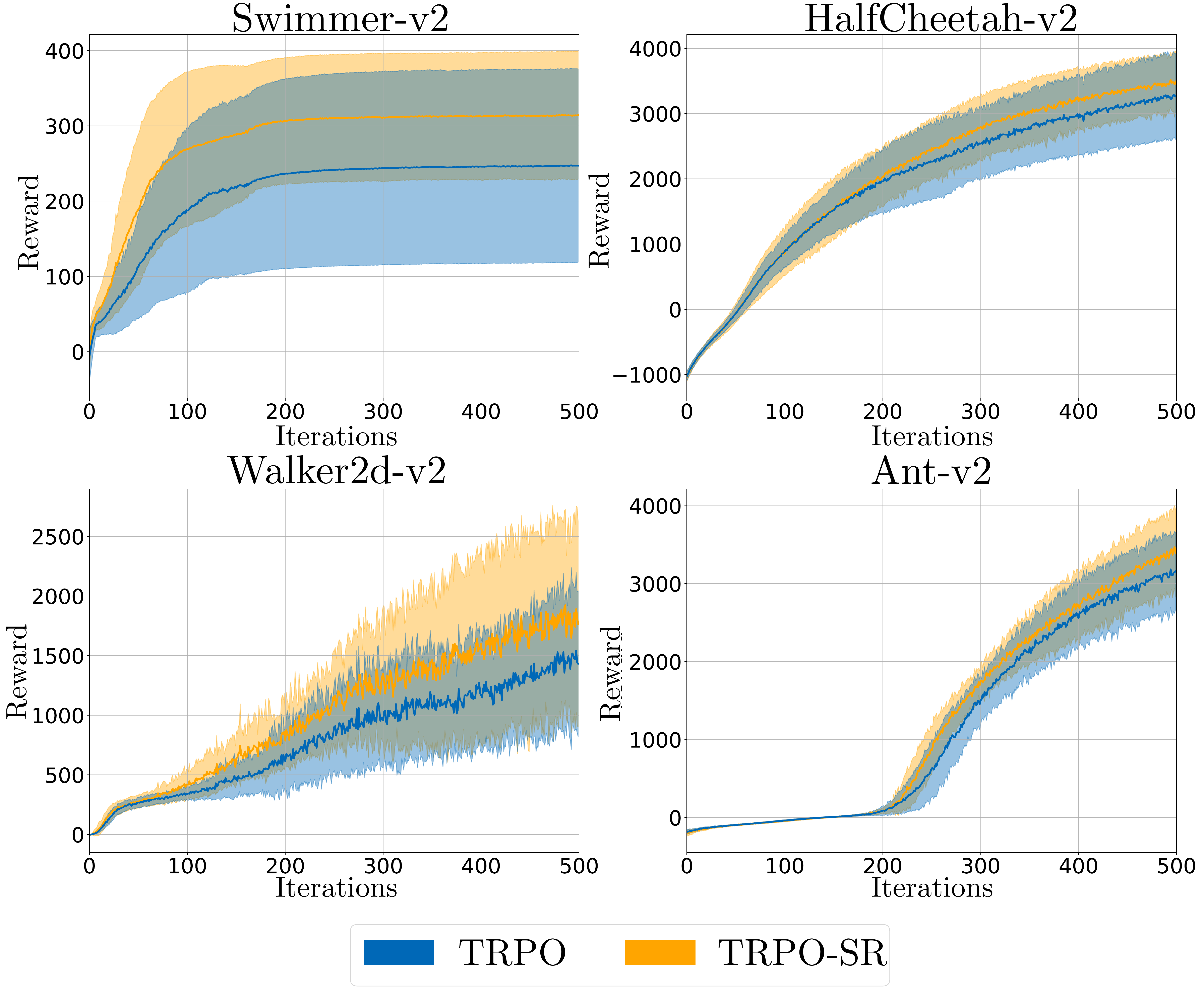}
		\includegraphics[width = 0.45\textwidth]{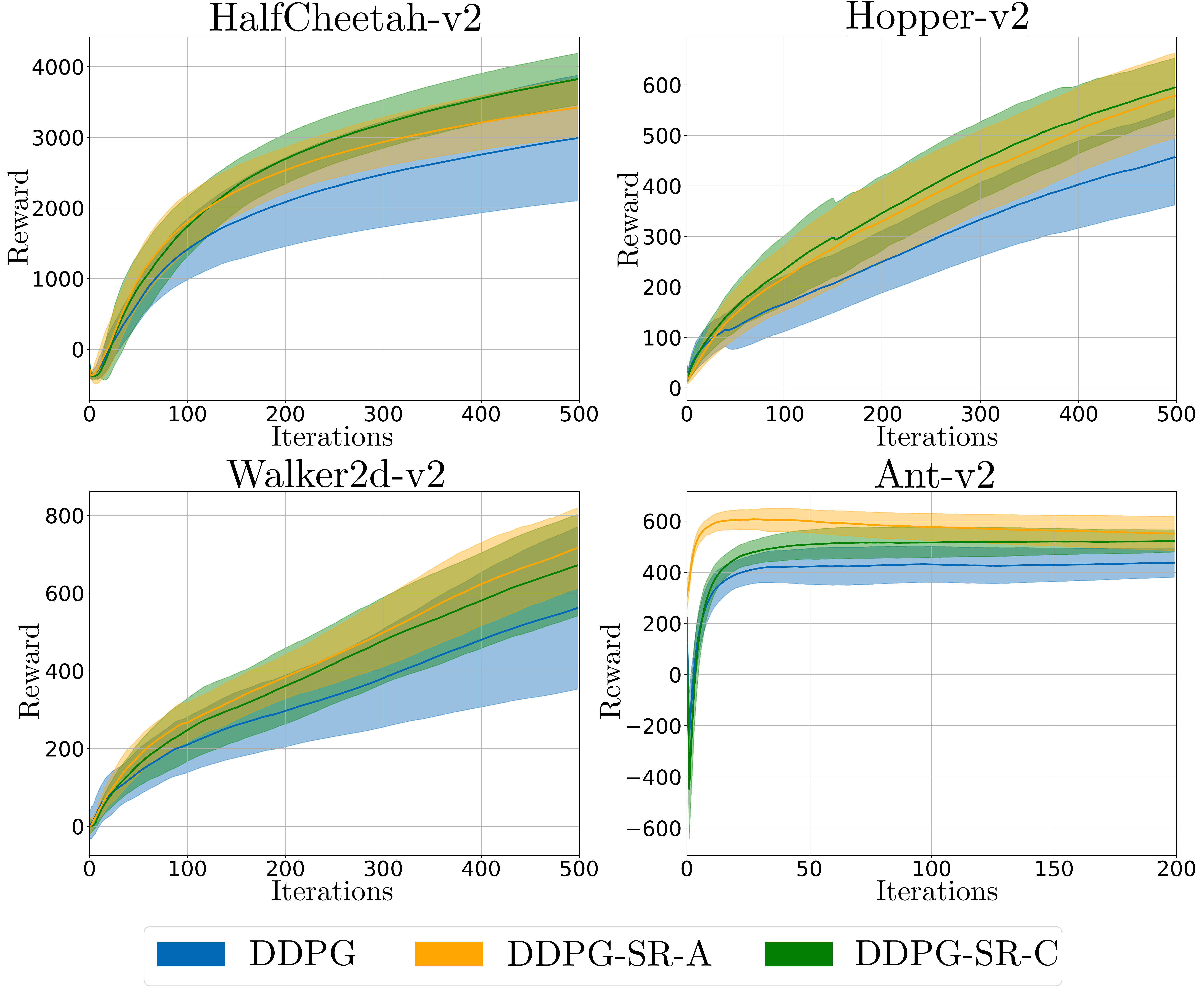}
	\vspace{-0.1in}
%	\caption{Learning curves (mean$\pm$standard deviation) for TRPO-SR (orange) trained policies versus the TRPO (blue) baseline when tested in clean environment. For all the tasks, TRPO-SR achieves a better mean reward than TRPO, with a reduction of variance in initial stage.
%	Learning curves (mean$\pm$standard deviation) for DDPG-SR-A (orange) and DDPG-SR-C (green) trained policies versus the DDPG (blue) baseline when tested in clean environment. For all the tasks, DDPG-SR-A(C) trained policies achieve better mean reward compared to  DDPG.}
	\caption{Learning curves (mean$\pm$standard deviation) for TRPO-SR and DDPG-SR compared to the baseline. For all the tasks, TRPO-SR achieves a better mean reward than TRPO, with a reduction of variance in initial stage. We observe similar phenomenon for DDPG-SR-A and DDPG-SR-C.}
	\vspace{-0.1in}
	\label{fig:srtrpo-learning-curve}
\end{figure*}

\section{Experiment}
We apply the proposed $\textbf{SR}^2\textbf{L}$ training framework to two popular reinforcement learning algorithms: TRPO and DDPG.
%Both of these algorithms have become the standard routine to  solve  large-scale control tasks, and the building blocks of many state-of-the-art reinforcement learning algorithms.
%For TRPO, we directly learns a smooth policy; For DDPG, we  promote the smoothness either in the actor (policy) or the critic (Q-function).
%
%
%
%
%%\subsection{Implementation}
%%\begin{figure*}[!ht]
%%	\centering
%%	\includegraphics[width = 0.95\textwidth]{./figure/gym.pdf}
%%	\vspace{-0.05in}
%%	\caption{OpenAI Gym Mujoco Benchmarks.  }
%%	\label{fig:gym}
%%\end{figure*}
Our implementation of $\textbf{SR}^2\textbf{L}$ training framework is based on the open source toolkit garage  \citep{garage}. We test our algorithms on OpenAI gym \citep{brockman2016openai} control environments with the MuJoCo \citep{todorov2012mujoco} physics simulator. For all tasks, we use a network of $2$ hidden layers, each containing $64$ neurons,  to parameterize the policy and the Q-function. 
For fair comparison, except for the hyper-parameters related to the smooth regularizer, we keep all the hyper-parameters the same as in the original implementation of garage.
We use the grid search  to select the hyper-parameters (perturbation strength $\epsilon$, regularization coefficient $\lambda_\mathrm{s}$) of the smoothness-inducing regularizer.
We set the search range to be $\epsilon \in [10^{-5}, 10^{-1}], \lambda_\mathrm{s} \in [10^{-2}, 10^{2}]$.
To solve the inner maximization problem in the update, we run $10$ steps of projected gradient ascent, with step size set as $0.2 \epsilon$.
For each algorithm and each environment, we train 10 policies with different initialization for 500 iterations (1K environment steps for each iteration).

\textbf{Evaluating the Learned Policies}.
%\textbf{TRPO with $\textbf{SR}^2\textbf{L}$} (TRPO-SR).
 We use Gaussian policy for TRPO in our implementation.
 Specifically, for a given state $s$,  the action follows a  Gaussian distribution $a \sim N(\pi_\theta(s), \sigma^2 I_A)$, where $\sigma$ is also a learnable parameter.
 Then the smoothness-inducing regularizer \eqref{eq:SR_stochastic_KLdiv} takes the form:
$ \cR_\mathrm{s}^\pi (\theta)  = \mathop{\mathbb{E}}_{s\sim\rho^\pi}\max_{\tilde{s}\in \mathbb{B}_d(s,\epsilon)} \norm{\pi_\theta(s) - \pi_\theta(\tilde{s})}_2^2/\sigma^2.
$ 

Figure \ref{fig:srtrpo-learning-curve} shows the mean and variance of the cumulative reward (over $10$ policies) for policies trained by TRPO-SR and TRPO for Swimmer, HalfCheetah, Hopper and Ant. For all the four tasks, TRPO-SR learns a better policy in terms of the mean cumulative reward. 
In addition, TRPO-SR enjoys a smaller variance of the cumulative reward with respect to different initializations.
These two observations confirm that our smoothness-inducing regularization improves sample efficiency as well as the training stability.

\begin{figure*}[!t]
	\centering
	\includegraphics[width = 0.45\textwidth]{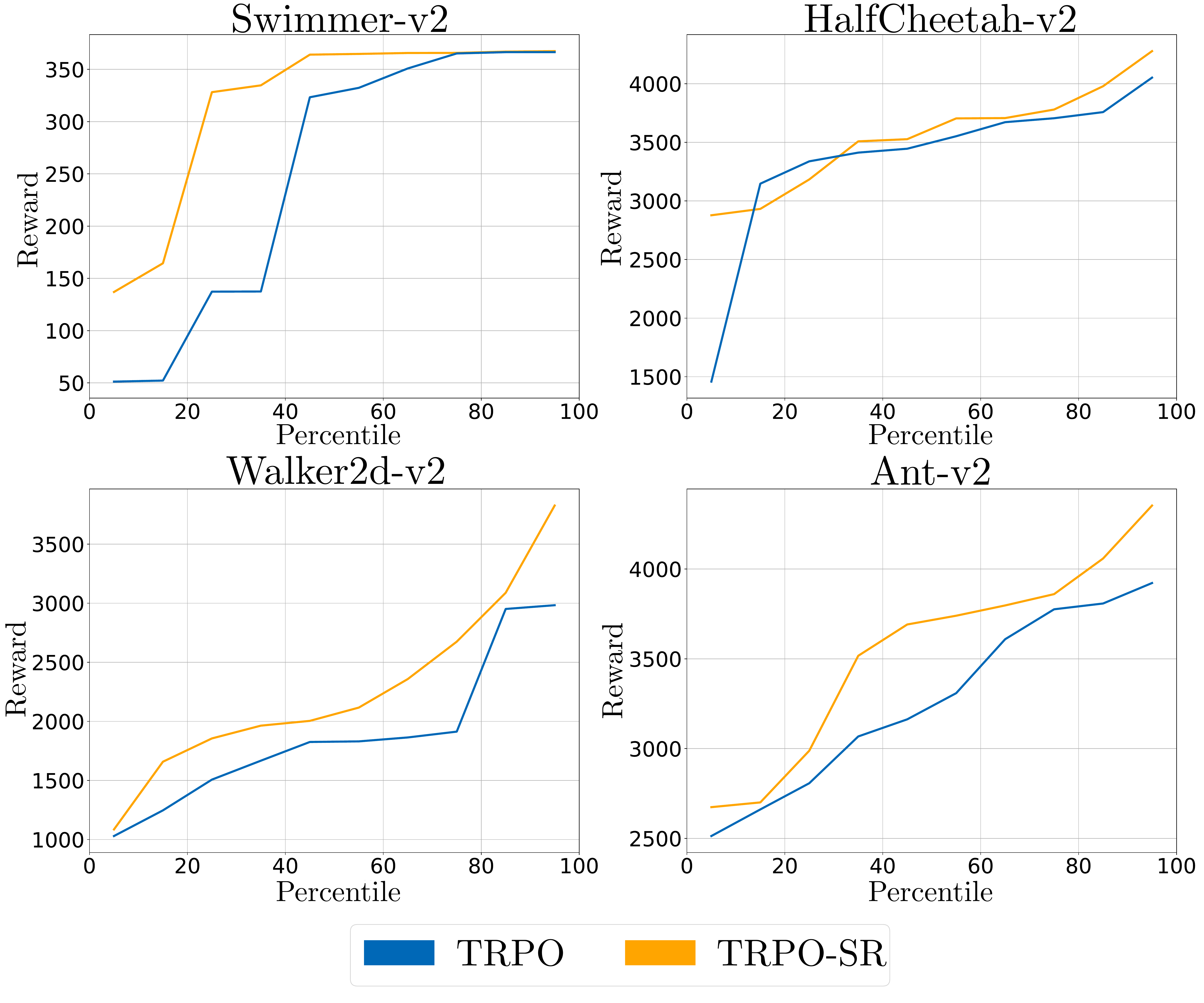}
		\includegraphics[width = 0.45\textwidth]{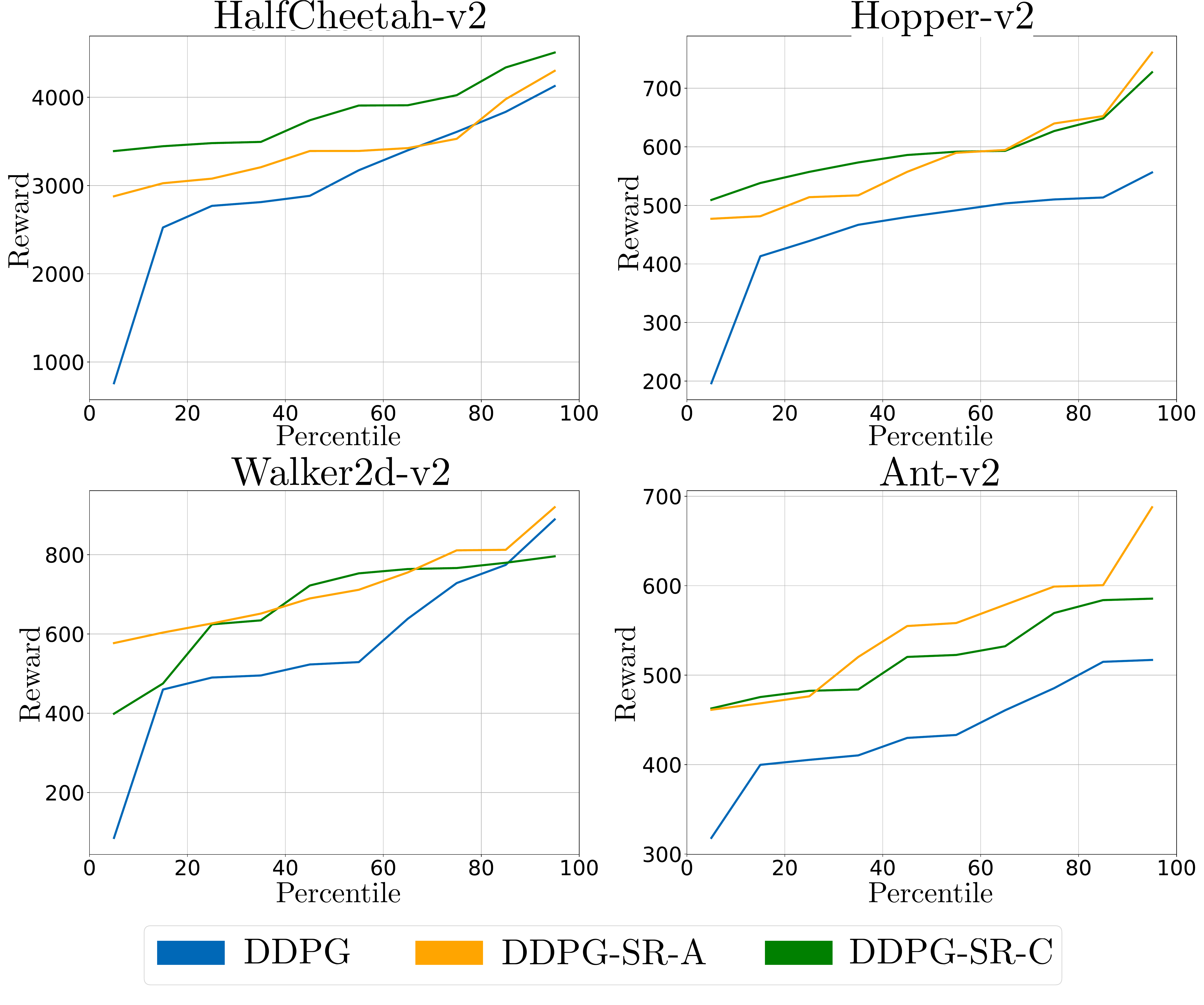}
	\vspace{-0.1in}
%	\caption{Percentile plots for TRPO-SR (orange) trained policies versus the TRPO (blue) baseline when tested in clean environment. Algorithms are run on ten different initializations and sorted to show the percentiles of cumulative reward. For all the tasks, TRPO-SR achieves better or competitive best performance,  and better worst case performance, compared with TRPO baseline.  
%	Percentile plots for DDPG-SR-A (orange) and DDPG-SR-C (green) trained policies versus the DDPG (blue) baseline when tested in clean environment. Algorithms are run with $10$ different initializations and sorted to show the  percentiles of cumulative reward. For all the tasks, DDPG-SR-A(C) achieves better or competitive best performance, and significantly better worst case performance, compared with DDPG baseline. }
	\caption{Percentile plots for TRPO-SR and DDPG-SR compared to the baseline. Algorithms are run on ten different initializations and sorted to show the percentiles of cumulative reward. For all the tasks, TRPO-SR achieves better or competitive best performance,  and better worst case performance, compared with TRPO baseline.  We observe similar behavior for DDPG-SR-A and DDPG-SR-C.
	 }
	 	\vspace{-0.1in}
	\label{fig:srtrpo-percentile-plot}
\end{figure*}

We further show that the advantage of our proposed $\textbf{SR}^2\textbf{L}$ training framework goes beyond improving the mean cumulative reward.
To show this, 
we run the algorithm with $10$ different initializations, sort the cumulative rewards of learned policies and plot the percentiles in Figure \ref{fig:srtrpo-percentile-plot}. For all four tasks, TRPO-SR uniformly outperforms the baseline TRPO.
For Swimmer  and HalfCheetach tasks, TRPO-SR significantly improves the worst case performance compared to TRPO, and have similar best case performance.
For Walker and Ant tasks,  TRPO-SR significantly improves the best case performance compared to TRPO.
We preform the same set of experiments on DDPG-SR compared to baseline DDPG, and observe the same behavior in Figure \ref{fig:srtrpo-learning-curve}.
%\begin{itemize}
%  \setlength\itemsep{-1.5 em}
%\item[$\bullet$] For all four tasks, TRPO-SR uniformly outperforms the baseline TRPO. \\
%\item[$\bullet$]For Swimmer  and HalfCheetach tasks, TRPO-SR significantly improves the worst case performance compared to TRPO, and have similar best case performance. \\
%\item[$\bullet$] For Walker and Ant tasks,  TRPO-SR significantly improves the best case performance compared to TRPO.
%\end{itemize}
Our empirical results show strong evidences that the proposed $\textbf{SR}^2\textbf{L}$ not only improves the average reward, but also makes the training process significantly more robust to failure case compared to the baseline method.

%\begin{figure}[!htb]
%	\centering
%	\includegraphics[width = 0.45\textwidth]{./figure/srddpg-learning-curve_1.pdf}
%	\vspace{-0in}
%	\caption{Learning curves (mean$\pm$standard deviation) for DDPG-SR-A (orange) and DDPG-SR-C (green) trained policies versus the DDPG (blue) baseline when tested in clean environment. For all the tasks, DDPG-SR-A(C) trained policies achieve better mean reward compared to  DDPG.}
%	\label{fig:srddpg-learning-curve}
%\end{figure}

%\textbf{DDPG with $\textbf{SR}^2\textbf{L}$}.
%We repeat the same evaluations for  applying the proposed $\textbf{SR}^2\textbf{L}$ framework to DDPG (DDPG-SR-A and DDPG-SR-C).
%Figure \ref{fig:srtrpo-learning-curve} shows the mean and variance of the cumulative reward for policies trained by DDPG-SR-A and DDPG-SR-C in HalfCheetah, Hopper and Walker2D and Ant environments. 
%For all the four tasks, DDPG-SR learns a better policy in terms of mean reward, which is consistent with our observations with TRPO-SR.
%For task Ant, DDPG-SR-A shows superior training stability, which is the only algorithm without drastic decay in the initial training stage.
%In addition, DDPG-SR-C shows competitive performance compared to DDPG-SR-A,
% significantly outperforms DDPG-SR-A and DDPG for task HalfCheetah.
% This shows that instead of directly learning a smooth policy, we can turn to learn a smooth Q-function and obtain similar performance benefits.  

Figure \ref{fig:srtrpo-percentile-plot}  plots percentiles of cumulative reward of learned policies using DDPG and DDPG-SR.
Similar to TRPO-SR, both DDPG-SR-A and DDPG-SR-C uniformly outperform the baseline DDPG for all the reward percentiles.
DDPG-SR is able to significantly improve the  the worst case performance, while maintaining competitive best case performance compared to DDPG.

%\begin{figure}[!htb]
%	\centering
%	\includegraphics[width = 0.45\textwidth]{./figure/srddpg-percentile-plot_1.pdf}
%	\vspace{-0in}
%	\caption{Percentile plots for DDPG-SR-A (orange) and DDPG-SR-C (green) trained policies versus the DDPG (blue) baseline when tested in clean environment. Algorithms are run with $10$ different initializations and sorted to show the  percentiles of cumulative reward. For all the tasks, DDPG-SR-A(C) achieves better or competitive best performance, and significantly better worst case performance, compared with DDPG baseline. }
%	\label{fig:srddpg-percentile-plot}
%\end{figure}

\begin{figure*}[!t]
	\centering
	\includegraphics[width = 0.48\textwidth]{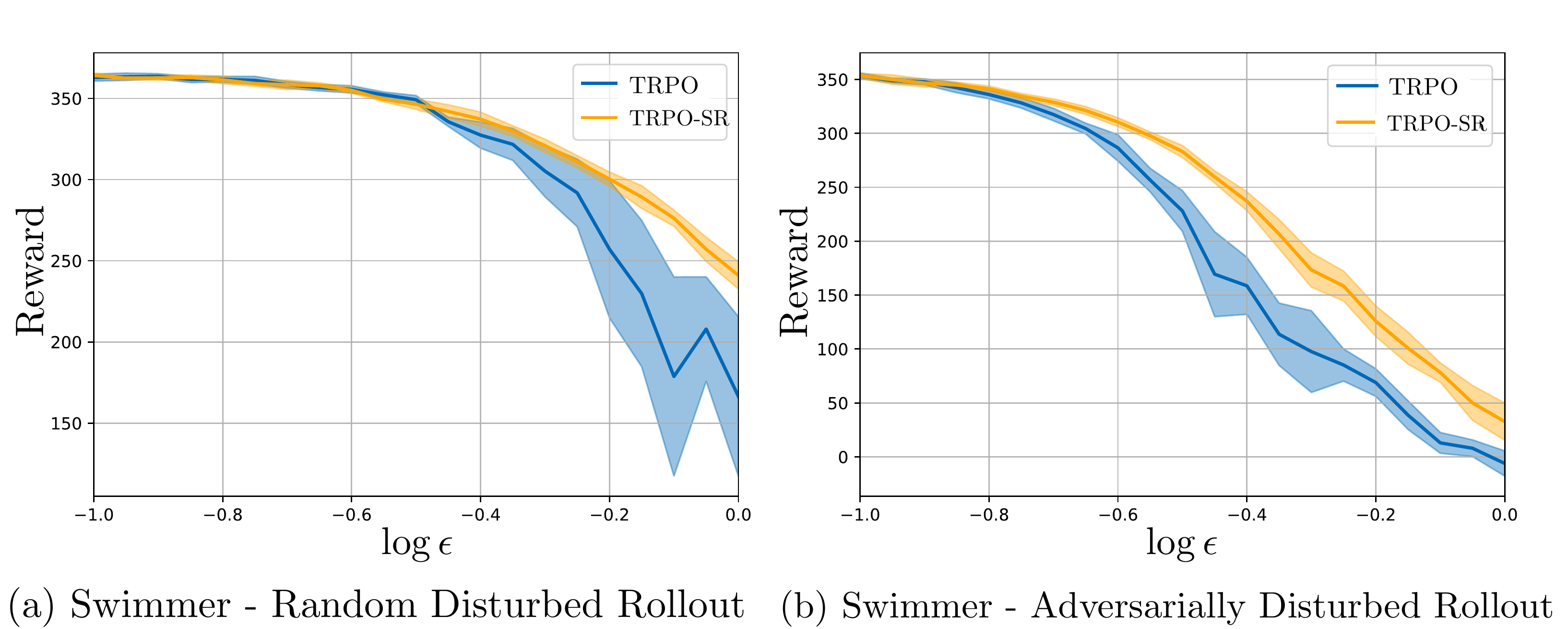}
		\includegraphics[width = 0.48\textwidth]{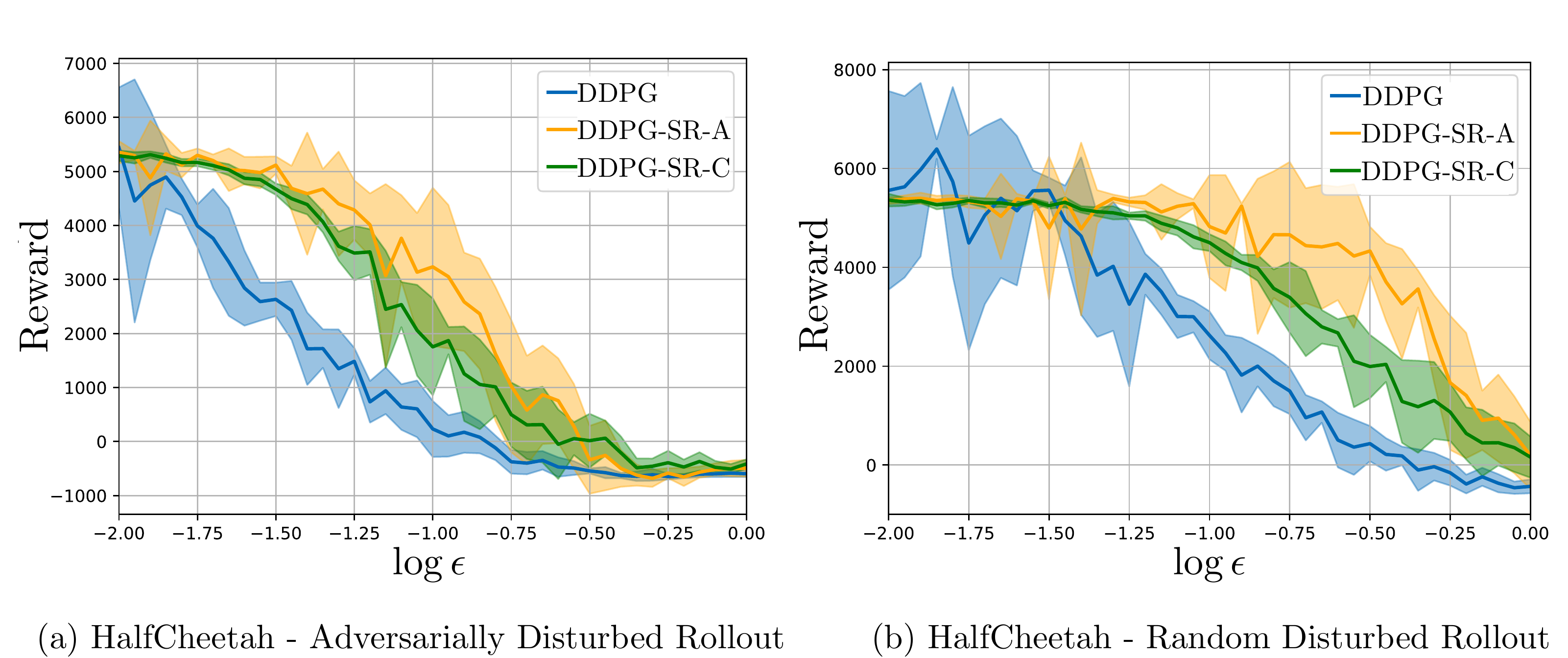}
%	\caption{Plot of cumulative reward (mean$\pm$standard deviation with multiple rollouts) of TRPO-SR (orange) trained policies versus the TRPO (blue) baseline when tested in disturbed environment. 
%	The random disturbance is uniformly sampled from set $\mathbb{B}_d(0, \epsilon) = \{\delta: \norm{\delta}_\infty \leq \epsilon\}$.
%	The adversarial disturbance belongs to the set $\mathbb{B}_d(0, \epsilon) = \{\delta: \norm{\delta}_\infty \leq \epsilon\}$.
%	TRPO-SR trained policies achieve a slower decline in performance than TRPO as we increase disturbance strength, and a significant reduction of variance under large disturbance strength.
%	Plot of cumulative reward (mean$\pm$standard deviation with multiple rollouts) of DDPG-SR-A (orange) and DDPG-SR-C (green) trained policies versus the DDPG (blue) baseline when tested in  disturbed environment. 
%	As  we increase disturbance strength $\epsilon$, DDPG-SR-A and DDPG-SR-C trained policies achieve a slower decline in performance than DDPG.
%	}
	\vspace{-0.1in}
	\caption{Plot of cumulative reward (mean$\pm$standard deviation with multiple rollouts) TRPO-SR and DDPG-SR compared to baseline, tested in disturbed environment. 
%	The random disturbance is uniformly sampled from set $\mathbb{B}_d(0, \epsilon) = \{\delta: \norm{\delta}_\infty \leq \epsilon\}$.
%	The adversarial disturbance belongs to the set $\mathbb{B}_d(0, \epsilon) = \{\delta: \norm{\delta}_\infty \leq \epsilon\}$.
	TRPO-SR trained policies achieve a slower decline in performance than TRPO as we increase disturbance strength, and a significant reduction of variance under large disturbance strength.
	We observe similar behavior for DDPG-SR-A and DDPG-SR-C.	}
	\vspace{-0.1in}
	\label{fig:trpo_robust}
\end{figure*}

\textbf{Robustness with Disturbance}.
We demonstrate that even if the $\textbf{SR}^2\textbf{L}$ training framework  is not targeting for robustness, the trained policy is still able to achieve robustness against both stochastic and adversarial measurement error, which is a classical setting considered in partially observable Markov decision process (POMDP) \citep{monahan1982state}. 
To show this, we evaluate the robustness of the proposed $\textbf{SR}^2\textbf{L}$ training framework in the Swimmer and HalfCheetah environments.
We evaluate  the trained policy with two types of disturbances in the test environment:
for a given state $s$, we add it with either 
(i)  random disturbance 
which are sampled uniformly from $\mathbb{B}_d(0,\epsilon)$,
or
 (ii) adversarial disturbance generated by solving:
$  \tilde{\delta} = \argmax_{\delta \in 
\mathbb{B}_d(0,\epsilon)}\cD(\pi_\theta(s),\pi_\theta(s + \delta ))
$ using $10$ steps of projected gradient ascent. 
For all evaluations, we use disturbance set $\mathbb{B}_d(0, \epsilon) = \{\delta: \norm{\delta}_\infty \leq \epsilon \}$.
For each policy and disturbed environment, we do $10$ stochastic rollouts to evaluate the policy and plot the cumulative reward of policy.

To evaluate the robustness of TRPO with $\textbf{SR}^2\textbf{L}$, we run both baseline TRPO and TRPO-SR in the Swimmer environment.
Figure \ref{fig:trpo_robust} plots the cumulative reward against the disturbance strength ($\epsilon$).
We see that for both random and adversarial disturbance, increasing the strength of the disturbance decreases the cumulative reward of the learned policies. 
On the other hand, we see that TRPO-SR clearly achieves improved robustness against perturbations, as its reward declines much slower than the baseline TRPO.
We see also see similar improvement of robustness on DDPG-SR-A and DDPG-SR-C in Figure \ref{fig:trpo_robust}.

	\vspace{-0.1in}
\section{Conclusion}
We develop a novel regularization based training framework $\textbf{SR}^2\textbf{L}$ to learn a smooth policy in reinforcement learning. 
The proposed regularizer encourages the learned policy to produce similar decisions for similar states. 
It can be applied to either induce smoothness in the policy directly, or induce smoothness in the Q-function, thus enjoys great applicability.
We demonstrate the effectiveness of $\textbf{SR}^2\textbf{L}$ by applying it to  two popular reinforcement learning algorithms, including TRPO and DDPG.
Our empirical results show that $\textbf{SR}^2\textbf{L}$ improves sample efficiency and training stability of current algorithms.
In addition, the induced smoothness in the learned policy also improves robustness against both random and adversarial perturbations to the state.

\newpage
\bibliographystyle{ims}
\bibliography{smoothrl}

\appendix
%!TEX root = ./SmoothRL.tex
\newpage 
\section{Appendix}
We present two variants of DDPG with the proposed smoothness-inducing regularizer. 
The first algorithm, DDPG-SR-A, directly learns a smooth policy with a regularizer that measures the non-smoothness in the actor network (policy).
The second variant, DDPG-SR-C, learns a smooth Q-function with a  regularizer that measure the non-smoothness in the critic network (Q-function). 
We present the details of DDPG-SR-A and DDPG-SR-C in Algorithm \ref{alg:ddpg-sr-a} and Algorithm \ref{alg:ddpg-sr-c}, respectively.

\begin{breakablealgorithm}
    \caption{ DDPG with smoothness-inducing regularization on the actor  (DDPG-SR-A).}
    \label{alg:ddpg-sr-a}
    \begin{algorithmic}
    \STATE{\textbf{Input}: step size for target networks $\alpha \in (0,1)$, coefficient of regularizer $\lambda_\mathrm{s}$, perturbation strength $\epsilon$, number of iterations to solve inner optimization problem $D$, number of training steps $T$, number of training episodes $M$, step size for  inner maximization $\eta_\delta$, step size for updating actor/critic network $\eta$. }
    \STATE{\textbf{Initialize}: randomly initialize the critic network $Q_\phi(s,a)$ and the actor network $\mu_\theta(s)$, initialize target networks $Q_{\phi'}(s,a)$ and $\mu_{\theta'}(s)$ with $\phi' = \phi$ and $\theta' = \theta$, initialize replay buffer $\cR$.}
    \FOR{ episode = 1 \ldots, M }
    \STATE{Initialize a random process $\epsilon$ for action exploration.}
    \STATE{Observe initial  state $s_1$.}
    \FOR{$t=1 \ldots T$}
    \STATE{Select action $a_t = \mu_{\theta}(s_t) + \epsilon_t$ where $\epsilon_t$ is the exploration noise.}
    \STATE{Take action $a_t$, receive reward $r_t$ and observe the new state $s_{t+1}$.}
    \STATE{Store transition $(s_t, a_t, r_t, s_{t+1})$ into the replay buffer $\cR$.}
    \STATE{Sample mini-batch $B$ of transitions $\{(s_t^i, a_t^i, r_t^i, s_{t+1}^i)\}_{i \in B}$ from the replay buffer $\cR$.}
    \STATE{Set $y_t^i = r_t^i + \gamma Q_{\phi'}(s_{t+1}^i, \mu_{\theta'}(s_{t+1}^i))$ for $i \in B$. }
    \STATE{Update the critic network: $\phi \leftarrow \argmin_{\tilde{\phi}} \sum_{i \in B} (y_t^i - Q_{\tilde{\phi}} (s_t^i, a_t^i))^2$.}
    \FOR{$s_t^i \in B$}
    \STATE{Randomly initialize $\delta_i$.}
    \FOR{$\ell = 1 \ldots D$}
    \STATE{$\delta_i \leftarrow \delta_i + \eta_\delta \nabla_\delta \norm{\mu_{\theta}(s_t^i) - \mu_{\theta}(s_t^i + \delta_i)  }_2^2$.}
    \STATE{$\delta_i \leftarrow \Pi_{\mathbb{B}_{d}(0,\epsilon)} (\delta_i)$.}
    \ENDFOR
       \STATE{Set $\hat{s}_t^i = s_t^i + \delta_i$.}
    \ENDFOR
    \STATE{Update the actor network: $$\theta \leftarrow \theta + \frac{\eta}{|B|} \sum_{i \in B} \rbr{ \nabla_a Q_\phi(s,a)\big|_{s= s_t^i, a = u_\theta(s_t^i)} \nabla_{\theta} \mu_\theta(s)\big|_{s = s_t^i}  - \lambda _s \nabla_\theta \norm{\mu_\theta(s_t^i) - \mu_\theta(\hat{s}_t^i)}_2^2}.$$}
    \STATE{Update the target networks: \begin{align*}
    \theta' &\leftarrow \alpha \theta + (1-\alpha) \theta', \\
    \phi' & \leftarrow \alpha \phi + (1-\alpha) \phi'.
    \end{align*}}
    \ENDFOR
    \ENDFOR
    \end{algorithmic}
\end{breakablealgorithm}

\begin{breakablealgorithm}
    \caption{DDPG with smoothness-inducing regularization on the critic  (DDPG-SR-C).}
    \label{alg:ddpg-sr-c}
    \begin{algorithmic}
    \STATE{\textbf{Input}: step size for target networks $\alpha \in (0,1)$, coefficient of regularizer $\lambda_\mathrm{s}$, perturbation strength $\epsilon$, number of iterations to solve inner optimization problem $D$, number of training steps $T$, number of training episodes $M$, step size for  inner maximization $\eta_\delta$, step size for updating actor/critic network $\eta$. }
    \STATE{\textbf{Initialize}: randomly initialize the critic network $Q_\phi(s,a)$ and the actor network $\mu_\theta(s)$, initialize target networks $Q_{\phi'}(s,a)$ and $\mu_{\theta'}(s)$ with $\phi' = \phi$ and $\theta' = \theta$, initialize replay buffer $\cR$.}    \FOR{ episode = 1 \ldots, M }
    \STATE{Initialize a random process $\epsilon$ for action exploration.}
    \STATE{Observe initial  state $s_1$.}
    \FOR{$t=1 \ldots T$}
    \STATE{Select action $a_t = \mu_{\theta}(s_t) + \epsilon_t$ where $\epsilon_t$ is the exploration noise.}
    \STATE{Take action $a_t$, receive reward $r_t$ and observe the new state $s_{t+1}$.}
    \STATE{Store transition $(s_t, a_t, r_t, s_{t+1})$ into replay buffer $\cR$.}
    \STATE{Sample mini-batch $B$ of transitions $\{(s_t^i, a_t^i, r_t^i, s_{t+1}^i)\}_{i \in B}$ from the replay buffer $\cR$.}
    \STATE{Set $y_t^i = r_t^i + \gamma Q_{\phi'}(s_{t+1}^i, \mu_{\theta'}(s_{t+1}^i))$ for $i \in B$. }
    \FOR{$s_t^i \in B$}
    \STATE{Randomly initialize $\delta_i$.}
    \FOR{$\ell = 1 \ldots D$}
    \STATE{$\delta_i \leftarrow \delta_i + \eta_\delta \nabla_\delta (Q_{\phi}(s_t^i,a_t^i)-Q_{\phi}(s_t^i + \delta,a_t^i))^2 $.}
    \STATE{$\delta_i \leftarrow \Pi_{\mathbb{B}_{d}(0,\epsilon)} (\delta_i)$.}
    \ENDFOR
       \STATE{Set $\hat{s}_t^i = s_t^i + \delta_i$.}
    \ENDFOR
    \STATE{Update the critic network: $$\phi \leftarrow \argmin_{\tilde{\phi}} \sum_{i \in B} (y_t^i - Q_{\tilde{\phi}} (s_t^i, a_t^i))^2 + \lambda_{\rm s}\sum_{i \in B} (Q_{\phi}(s_t^i,a_t^i)-Q_{\phi}(\hat{s}_t^i,a_t^i))^2.$$}
    \STATE{Update the actor network: $$\theta \leftarrow \theta + \frac{\eta}{|B|} \sum_{i \in B}  \nabla_a Q_\phi(s,a)\big|_{s= s_t^i, a = u_\theta(s_t^i)} \nabla_{\theta} \mu_\theta(s)\big|_{s = s_t^i} .$$}
    \STATE{Update the target networks: \begin{align*}
    \theta' &\leftarrow \alpha \theta + (1-\alpha) \theta', \\
    \phi' & \leftarrow \alpha \phi + (1-\alpha) \phi'.
    \end{align*}}
    \ENDFOR
    \ENDFOR
    \end{algorithmic}
\end{breakablealgorithm}

\end{document}